\pdfoutput=1

\documentclass[11pt]{article}

\usepackage{ACL2023}

\usepackage{times}
\usepackage{latexsym}

\usepackage[T1]{fontenc}

\usepackage[utf8]{inputenc}

\usepackage{microtype}

\usepackage{inconsolata}

\usepackage{amsmath, amsthm, amssymb, amsfonts}
\usepackage{algorithm}
\usepackage{booktabs}
\usepackage{multicol}
\usepackage{multirow}
\usepackage{enumitem}
\usepackage{booktabs}
\usepackage{graphicx}
\usepackage{tabularx}
\usepackage{xcolor}
\usepackage[noend]{algpseudocode}
\usepackage[subtle]{savetrees}
\newcommand{\name}{\textsc{TAROT}}

\title{\name : Task-Oriented Authorship Obfuscation Using Policy\\ Optimization Methods}

\author{Gabriel Loiseau$^{1,2}$ \quad Damien Sileo$^{2}$ \quad Damien Riquet$^{1}$ \quad Maxime Meyer$^{1}$ \quad Marc Tommasi$^{2}$ \\ $^{1}$Hornetsecurity, Hem, France \\ $^2$Univ. Lille, Inria, CNRS, Centrale Lille, UMR 9189 - CRIStAL, F-59000 Lille, France \\
\texttt{gabriel.loiseau@inria.fr}}

\begin{document}
\maketitle
\begin{abstract}
Authorship obfuscation aims to disguise the identity of an author within a text by altering the writing style, vocabulary, syntax, and other linguistic features associated with the text author. 
This alteration needs to balance privacy and utility. While strong obfuscation techniques can effectively hide the author's identity, they often degrade the quality and usefulness of the text for its intended purpose. Conversely, maintaining high utility tends to provide insufficient privacy, making it easier for an adversary to de-anonymize the author. Thus, achieving an optimal trade-off between these two conflicting objectives is crucial.
In this paper, we propose 
\name: \textbf{T}ask-Oriented \textbf{A}utho\textbf{r}ship \textbf{O}bfuscation Using Policy Op\textbf{t}imization,
a new unsupervised authorship obfuscation method whose goal is to optimize the privacy-utility trade-off by regenerating the entire text considering its downstream utility. Our approach leverages policy optimization as a fine-tuning paradigm over small language models in order to rewrite texts by preserving author identity and downstream task utility. We show that our approach largely reduces the accuracy of attackers while preserving utility. We make our code and models publicly available.\footnote{\url{https://github.com/hornetsecurity/tarot}}

\end{abstract}

\section{Introduction}

Text is a primary medium for storing user data, training machine learning models, and interacting with large language models (LLMs) during inference. However, it also poses significant privacy risks, as sensitive or personal information contained within text can be exposed or misused. Text anonymization is a vital technique to address these concerns by removing or obfuscating personal information. This process protects individual privacy while ensuring that machine learning models can still derive meaningful insights and patterns from anonymized data, preserving its utility.

\begin{figure}
    \centering
    \includegraphics[width=1\columnwidth]{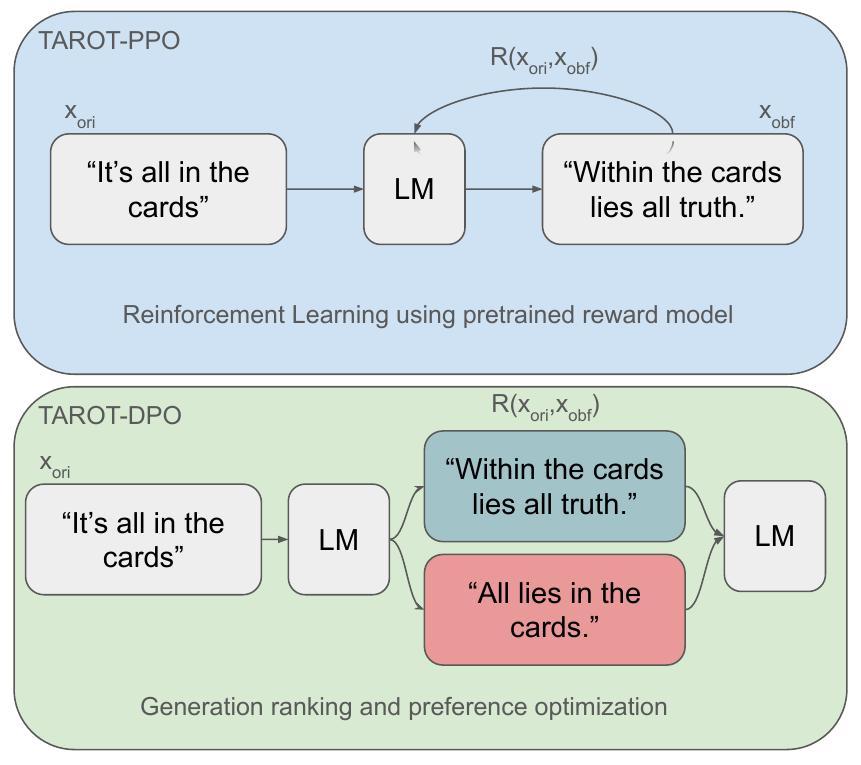}
    \caption{Illustration of the two versions of \name: We generate obfuscation candidates and optimize the best policy using reinforcement learning and preference optimization.}
    \label{fig:tarot-detail}
    \vspace{-4mm}
\end{figure}

Currently, most work done on text anonymization focuses on redacting sensitive entities in a given document \cite{lison-etal-2021-anonymisation}. This is sufficient for texts where the only private aspects are named entities, such as medical reports, court cases, or biographies. But it is inadequate for removing the author's writing style, or the weak signals that can be used as hints for identification, which is, for example, the case for blog articles or emails. Redacting entities in text while keeping stylometric features linked to a specific individual would eventually result in a leak of information. Indeed, the writing style is a strong indicator of a person’s identity \cite{MostellerWallace}. Previous work on authorship attribution highlights the large amount of information that can be extracted from seemingly anonymized texts and the ease of identification of authors, especially for long documents \cite{fabien-etal-2020-bertaa}. 

To solve this issue, authorship obfuscation (AO) aims to hide the author's identity by replacing some part of the text associated with authorship indicators. Modifying the original text can impact its usability for specific tasks (i.e. utility), and therefore badly affects the downstream performances and text comprehension of machine learning models. The enforcement of privacy creates a trade-off between privacy and utility, where keeping the original text preserves the unchanged utility of the text, while not defending against attribution attacks. On the other hand, obfuscating the entire text guarantees privacy, but leads to unusable text in practice.
Previous approaches design their obfuscation by maximizing the preserved text content. They limit the modifications to small and targeted edits in order to preserve text meaning and keep textual content as close as possible to the original. While this strategy is necessary to maintain the exact content and ensure that we convey the exact same message (before publishing the text online for example), those approaches often lead to insufficient modification in the text, especially against realistic attack scenarios 
\cite{zhai-etal-2022-adversarial}.

To address these limitations, we reframe the AO problem into an adversarial problem between two adversaries (e.g. machine learning models): one attacker model whose goal is to reveal the identity of a given author from written texts, and one utility model that aims to perform a given task using authors' data. The goal is to provide a modified version of the original text such that the utility model can accurately perform its task while preventing the attacker from identifying the author, making the obfuscation task-oriented. This perspective is more angled towards data users who need to privately perform utility tasks on the data, where some degree of content alteration may be acceptable if it enhances privacy.
The notion of task-oriented obfuscation/anonymization also takes its origin in the law. As stated by GDPR \cite{GDPR2016a}, the collection and processing of personal information (including written texts) must be specified for a given usage.

In order to learn this privacy-utility trade-off, we use the combination of supervised fine-tuning (SFT) and policy optimization (PO) to guide a generative model into generating privacy- and utility-preserving outputs. Our model learns to rewrite the text while removing potential authorship signals, and preserving the text utility for a downstream task. This rewriting goal is further validated by the conclusion of \citet{10.1093/idpl/ipac008} which showed how difficult it is to comply with GDPR requirements concerning text anonymization without changing the entire text. 

We fine-tune a text simplification model for AO using a customized reward model. We design an unsupervised reward model for PO using two pretrained sentence
embedding models. The utility reward penalizes the fact that the General Text Embeddings \cite{li2023gte} of the anonymized sentence is too far removed from that of the original sentence. The author rewards does the opposite on the embedding built by the  Universal Authorship Representation model from \citet{rivera-soto-etal-2021-learning}.
Our final models are trained in an open-world setting where the number of authors is not defined, the same goes for the end utility for our model to work on a multi-task setting. We also provide experimentation on three different datasets, movie reviews, blog articles and scholar documents. We show that \name~can be used on multiple datasets targeting different tasks while protecting authorship.

In summary, we list the main contributions as follows:
\begin{itemize}
    \item We design a new framework for task-oriented AO by leveraging PO algorithms to maximize the end usage of data. The objective is to help reduce the traditional constraints associated with utility preservation in the literature (strict content preservation and semantic quality) by looking for a downstream classification task to achieve with the anonymized data. 
    \item Starting from this framework, we propose \name, a task-oriented generation model aiming to obfuscate text without any prior knowledge of the author (making it unsupervised, and usable on any dataset, even if the authors are not clearly indicated) while maximizing the utility for a variety of tasks. We release two versions of \name~from two different fine-tuning PO algorithms: \name-PPO and \name-DPO.
    \item We further evaluate \name~on three datasets associated with different classification tasks, using different authorship attackers and downstream usage scenarios.
\end{itemize}

\section{Related Work}
\paragraph{Authorship Obfuscation} 
Obfuscation techniques can be regrouped into two categories, depending on their implementation. Generic methods, on one hand, are methods that were not explicitly designed for AO, but show interesting performance. These methods include machine translation \cite{altakrori-etal-2022-multifaceted, KeswaniT0M16}, paraphrasing \cite{krishna2023paraphrasing}, or synonym replacements \cite{Potthast2016AuthorOA}. 

More recently, advanced techniques were built explicitly for AO, often relying on a trained attacker performing authorship attribution attacks on the obfuscated text. Then, they perform accurate adversarial text edits from the attacker knowledge on authors in order to obtain a privatized output. Mutant-X \cite{mahmood2019mutantx}, is a genetic algorithm that utilizes GloVE \cite{pennington-etal-2014-glove} word embeddings selected from an SVM or Random Forest attacker to replace words in a document with similar ones. 

Jamdec \cite{fisher2024jamdec} is an unsupervised approach for obfuscating the writing style of text while preserving semantics. It uses embedding-based and likelihood-based methods, rather than attacker-based methods, to extract keywords, then generates multiple text variations using Constrained Diverse Beam Search on GPT2-XL (1.61B parameters). Finally, the candidates are filtered using Natural Language Inference (NLI) and Corpus of Linguistic Acceptability (CoLA) metrics to ensure coherence, content preservation, and grammatical correctness.

Recently, ALISON \cite{xing2024alison} employs a lightweight multilayer perceptron classifier using part-of-speech sequences to guide obfuscation, and leverages a BERT pre-trained language model to generate replacement sequences. By ranking and replacing important part-of-speech n-grams, ALISON obfuscates text uniformly, reducing classifier confidence.

Related studies share a common approach to evaluating privacy: they measure it through the performance of authorship attribution classifiers against obfuscated texts. \citet{zhai-etal-2022-adversarial} push forward this evaluation framework by introducing adversarial attackers that can resist obfuscation techniques. For measuring utility, the standard is to treat AO as a reference-less natural language generation problem, and to rely on standard metrics used for similar tasks such as machine translation and summarization \cite{altakrori-etal-2022-multifaceted}.

\paragraph{Reinforcement Learning}
In NLP, reinforcement learning (RL) is often used to capture small signals over word or sentence embedding. For example, \citet{mosallanezhad-etal-2019-deep} proposes a text representation anonymization approach that employs deep reinforcement learning to detect and modify text embeddings to maintain a good privacy-utility trade-off.  

With the development of Reinforcement Learning from Human Feedback (RLHF) as a LLM fine-tuning paradigm, RL techniques have been leveraged to improve language models with scalar metrics by optimizing rewards from (human) feedback. It has emerged as a prominent tool for tackling undesirable behaviors such as toxicity, social biases, and offensive language \cite{ouyang2022training}. This is accomplished by implementing PO algorithms to optimize a language model (LM) by associating a reward with each generation, derived from a trained reward model. 

Very recently, \citet{liu2024authorship} introduced an authorship style transfer method using PO. They optimize style transfer generation using style similarity reward models. Authorship style transfer is similar to AO in the way those task's goal is to change within a text the author writing style. However, style transfer assumes a distinct target style to achieve, whereas AO assumes a lack of distinct style. \citet{fisher2024jamdec} also showed the ineffectiveness of style transfer for AO. To the best of our knowledge, our work is the first one applying PO algorithms on AO.

\paragraph{Private Synthetic Text Generation}
Our work lies at the frontier between private text editing and synthetic text generation. Creating private synthetic data often relies on established frameworks such as differential privacy \cite{dp}.
In contrast to these approaches, we focus on the implementation of a single text-to-text transformation specifically designed for authorship obfuscation, rather than on the generation of new textual data derived from potentially multiple sources \cite{mattern-etal-2022-differentially}.

Differential privacy traditionally targets noise addition in documents to produce useful and private text representations \cite{feyisetan2019leveraging,Fernandes}.
Applying differential privacy to document rewriting primarily serves to mitigate membership inference attacks, addressing a distinct threat model compared to the authorship attribution attacks targeted by our approach. While these techniques exhibit emergent capabilities for masking authorship signals \cite{dp-bart, dp-vae, dp-prompt}, they typically do so at a substantial cost to text utility, both at the task-level and the syntactic-level \cite{mattern-etal-2022-limits}. This approach introduces unnecessary noise to semantic content not relevant to authorship identification, often degrading the overall coherence and readability of the text. In contrast, our obfuscation methodology implements targeted modifications to stylometric features while maintaining the overall integrity of the source text.

\section{Methodology}
\subsection{Problem Formulation}
Let $x_{\text{ori}}$ represent the original document authored by a specific author $a \in \mathcal{A}$.  $\mathcal{A}$ denoting a predetermined set of authors.  The objective of authorship obfuscation is to generate a new document, denoted as $x_{\text{obf}}$, which cannot be attributed to the original author $a$. To assess the effectiveness of obfuscation, we employ a classification model, denoted as $f_{attr}(\cdot)$ (i.e. an authorship attribution model), which has been trained to distinguish documents based on their respective authors within $\mathcal{A}$. The goal of authorship obfuscation is to design an obfuscation method $O(\cdot)$, such that $f_{attr}(O(x_{\text{ori}})) \neq f_{attr}(x_{\text{ori}}).$

In addition, a successful obfuscation algorithm would not only trick an attacker into predicting the wrong author, but also preserve the document utility for downstream usage. In this paper, instead of mainly measuring this utility change though various semantic or content preservation metrics (i.e. METEOR score, BERT score, etc.) we highlight the selection of a prior task $\mathcal{T}$ in order to evaluate obfuscation with respect to $\mathcal{T}$. We denote as $f_{\mathcal{T}}(\cdot)$ the classification model used for a utility task.
An ideal $O(\cdot)$ would preserve the original label $f_{\mathcal{T}}(O(x_{\text{ori}})) = f_{\mathcal{T}}(x_{\text{ori}})$. 

Note that $\mathcal{T}$ is likely not known when we train the obfuscation model, underscoring the necessity for a versatile obfuscation strategy. This task-agnostic approach prevents the obfuscation model from learning to transform the text specifically to fit the label of $\mathcal{T}$, which would compromise its generality across different tasks.

\subsection{Framework Overview}
Our task-oriented framework can be decomposed in two steps. First, we initialize our generation model from a SFT baseline, this will first guide our LM to generate modified versions of the input text instead of proceeding text copy. Second, we apply a PO algorithm to fine-tune our SFT model. We experiment with two different PO algorithms, Proximal Policy Optimization \cite{schulman2017ppo} and Direct Preference Optimization \cite{rafailov2023dpo} (see Figure~\ref{fig:tarot-detail}). We optimize our SFT generations using a reward model composed of both privacy and content preservation components.

\subsection{SFT Initialization}
First, we use a fine-tuned LM to initiate our text generation task. 
We employ the \textit{Keep It Simple}\footnote{\url{https://hf.co/philippelaban/keep_it_simple}} simplification model \cite{laban-etal-2021-keep} as an SFT baseline. This model is a fine-tuned version of \texttt{GPT2-medium} on the Newsela\footnote{\url{https://newsela.com/}} dataset for text simplification.
The utilization of a simplification model encourages a reduction in the amount of information conveyed by a sentence, thereby affording the opportunity to eliminate author-specific features\footnote{Our preliminary experiments revealed that using a simplification model outperformed comparable models of similar size for copy, paraphrasing, back-translation, and summarization, delivering superior privacy and utility.}. To our knowledge, this is the first time that a simplification model has been used for AO.
Moreover, our framework is broadly compatible with any autoregressive LM, and can be adapted with larger architectures and other generation tasks. 

\begin{table*}[ht]
\centering
\resizebox{\linewidth}{!}{
\begin{tabular}{lcccccc}
\toprule
\textbf{Dataset} & \textbf{Authors} & \textbf{Texts} & \textbf{\begin{tabular}[c]{@{}c@{}}Avg. Texts / Author\\ (std)\end{tabular}} & \textbf{\begin{tabular}[c]{@{}c@{}}Avg. Words / Text\\ (std)\end{tabular}} & \textbf{\begin{tabular}[c]{@{}c@{}}Avg. Tokens / Text\\ (std)\end{tabular}} & \textbf{\begin{tabular}[c]{@{}c@{}}Avg. Chars / Text\\ (std)\end{tabular}}  \\ 
\midrule
\multirow[c]{2}{*}{IMDb} & $10$ & $10000$ & $1000 (\pm 0)$ & $364 (\pm 209)$ & $393 (\pm 228)$ & $1869 (\pm 1077)$ \\
    & $20$ & $20000$ & $1000 (\pm 0)$ & $345 (\pm 209)$ & $371 (\pm 225)$ & $1767 (\pm 1081)$ \\
\midrule
\multirow[c]{2}{*}{BAC} & $10$ & $23534$ & $2353 (\pm 639)$ & $118 (\pm 195)$ & $120 (\pm 236)$ & $524 (\pm 1027)$ \\
    & $20$ & $39379$ & $1969 (\pm 599)$ & $118 (\pm 175)$ & $123 (\pm 214)$ & $529 (\pm 921)$ \\
\midrule
\multirow[c]{2}{*}{AMT} & $10$ & $196$ & $20 (\pm 2)$ & $497 (\pm 14)$ & $592 (\pm 41)$ & $2956 (\pm 194)$ \\
    & $20$ & $362$ & $18 (\pm 2)$ & $502 (\pm 102)$ & $590 (\pm 38)$ & $2956 (\pm 207)$ \\
\bottomrule
\end{tabular}}
\caption{Dataset statistics}
\label{tab:dataset_statistics}
\end{table*}

\subsection{Policy Optimization Algorithms}
We use two different PO algorithms to optimize generations of our SFT baseline.
The Proximal Policy Optimization (PPO) \cite{schulman2017ppo} algorithm is a policy gradient method whose goal is to optimize a policy  with respect to continuous rewards. In our case, a policy is a generation strategy, i.e. a final LM.
Initialized from the SFT policy, we sample completions $y$ given prompts $x$ and the reward model parametrized by $\phi$ produces a score $r_\phi(x,y)$ based on these completions. The reward score $r_\phi(x,y)$ is then combined with a Kullback–Leibler (KL) penalty to ensure the policy does not deviate too much from the SFT policy (leading to unusable generations). Specifically, the reward of the RL problem is:
\begin{equation*}\label{eq:RL}
\resizebox{\columnwidth}{!}{
$R(x, y) = r_{\phi}(x, y) - \beta\mathbb{D}_{\textrm{KL}}\bigl[\pi_{\theta}(y\mid x)\mid \mid \pi_\text{SFT}(y\mid x)\bigr]$}
\end{equation*}

where $\beta$ is a parameter controlling the strength of the KL penalty, $\theta$ the parameters of RL policy $\pi_{\theta}$, and $r_{\phi}$ the reward model with parameters $\phi$. Then, PPO is used to maximize the following objective:  $$\max_{\pi_{\theta}}\   \mathbb{E}_{x\sim \mathcal{D}_\text{SFT}, y\sim \pi_{\theta}(y \mid x)} R(x, y)$$ where $\mathcal{D}_\text{SFT}$ is the prompts in the SFT dataset. 

\citet{rafailov2023dpo} later introduced the Direct Preference Optimization
(DPO) algorithm, which implicitly optimizes the same objective as PPO. DPO directly optimizes the model by a straightforward contrastive loss, boosting the reward of the preferred generation $y_c$ and penalizing the one of the non-preferred generation $y_r$ from a prompt $x$. DPO is a RL-free approach which has the following loss:
\begin{equation*}
    -\log \sigma \left(\beta \log \frac{\pi_{\theta}(y_c\mid x)}{\pi_{\text{SFT}}(y_c\mid x)} - \beta \log \frac{\pi_{\theta}(y_r\mid x)}{\pi_{\text{SFT}}(y_r\mid x)}\right)
\end{equation*} 

where $\sigma$ is the sigmoid function, and $\beta$ the scaling parameter.
In this study, we lack access to a preference dataset for DPO fine-tuning. Consequently, following the methodology of \citet{rafailov2023dpo}, we generate this dataset by sampling responses from the same SFT dataset, and we rank those preferences using the same reward model (see Appendix~\ref{app:dpo-training}). This is justified as it is not possible to obtain a preference dataset from human feedback in the AO setting.


\section{Experimental Setup}
In this section, we describe the datasets involved for training and evaluation of our resulting models, and present our custom reward targeting the open-world authorship verification and multi-task text embeddings to learn this AO task. We then evaluate the resulting obfuscation against text edition and rewriting baselines.

\subsection{Datasets}
\paragraph{Training} We use a separate dataset to train our PO models. We fine-tune our base simplification model on the Yelp reviews dataset\footnote{\url{https://hf.co/datasets/yelp_review_full}} \cite{yelp_NIPS2015} composed of reviews from Yelp. The dataset is extracted from the Yelp Dataset Challenge 2015. This dataset is employed in an unsupervised way, to ensure we train our models on a large number of authors.

\paragraph{Evaluation} To evaluate our obfuscation models, we use three different datasets. (i) IMDb62\footnote{\url{https://hf.co/datasets/tasksource/imdb62}}, is a subset of the IMDb Authorship Attribution dataset initially presented by \citet{seroussi2014IMDb}. It consists of 62 authors with 1,000 texts per author taken from IMDb movie reviews. The utility task associated with this dataset is the review sentiment. For this, we map the movie rating between 0 and 10 associated with each review to a sentiment between \emph{positive} and \emph{negative}. A positive review occurs when the review rating is strictly larger than 5. (ii) The Blog Authorship Corpus\footnote{\url{https://u.cs.biu.ac.il/~koppel/BlogCorpus.htm}} dataset \cite{Schler2006BAC}  consists of aggregated blog posts from 19,320 bloggers gathered from blogger.com. We pick the list of 13 topics present in the dataset as the utility task. (iii) The Extended-Brennan-Greenstadt\footnote{\url{https://hf.co/datasets/tasksource/Drexel-AMT}} dataset \cite{amtdataset} is composed of short paragraphs about scholar subjects gathered from 42 different authors from Amazon Mechanical Turk. The utility task of this dataset is indicated by the ``background'' column, as a binary classification problem.

For all datasets, we create two subsets containing the texts from 10 and 20 authors. For the Blog Authorship Corpus, we select the authors with the highest number of texts. We select the 10 (resp. 20) first authors listed in IMDb62 and Extended-Brennan-Greenstadt. We report summary statistics of each dataset in Table~\ref{tab:dataset_statistics} and refer to every dataset as IMDb, BAC, and AMT followed by the number of considered authors. In summary, IMDb has rather long texts, numerous texts per author with a large associated standard deviation. BAC texts are shorter, with a higher number of texts per author compared to IMDb. Finally, for the AMT dataset, the texts are the longest with few variations, and the number of texts per author is the smallest.

\subsection{Reward Models}
To perform PO, we build a reward model from two different rewards components targeting respectively text semantics and text authorship, aiming to disentangle privacy and utility to control the trade-off. 

For utility, we use a pretrained General Text Embeddings (GTE) \cite{li2023gte} to represent the reward as a cosine similarity between GTE before and after obfuscation\footnote{We use the \texttt{gte-large-en-v1.5} from \texttt{sentence-transformers} \url{https://hf.co/Alibaba-NLP/gte-large-en-v1.5}}. Denote as $\mathrm{GTE}(x)$ the embedding vector of size 1024, our utility reward is defined as:
\begin{equation*}
    R_{util} = \mathrm{cossim}(\mathrm{GTE}(x_{ori}),\mathrm{GTE}(x_{obf}))
\end{equation*}

For the privacy reward, we use the Learning Universal Authorship Representations model (LUAR), from \citet{rivera-soto-etal-2021-learning}. LUAR's goal is to transform a given text into a 512 dimensions embedding, such that representations of texts by the same author are closer, according to cosine similarity, than those by other authors.

Denote as $\mathrm{LUAR}(x)$ the embedding vector given by the LUAR model, our privacy reward is defined as:
\begin{equation*}
    R_{priv} = 1 - \mathrm{cossim}(\mathrm{LUAR}(x_{ori}),\mathrm{LUAR}(x_{obf}))
  \end{equation*}
where $\mathrm{cossim}$ denotes the cosine similarity. 

We obtain our final reward by summing the two previous rewards $R = R_{util} + R_{priv}$. All implementation details are listed in Appendix~\ref{apx:hardware-code}.

\subsection{Evaluation}
\paragraph{Privacy Metrics}
The goal for obfuscation is to change the text in order to reduce as much as possible the attacker accuracy. We employ authorship attribution as an evaluation attacker to simulate an attack scenario when the attacker has already access to some sample data of targeted authors to train an attacker classifier. This is a stronger scenario than directly using the reward model as evaluation, since it only assumes one-to-one comparison between texts. For each evaluation dataset, we train a DeBERTa-v3 \cite{he2021deberta} model as an authorship attribution classifier. We split each evaluation dataset in 80\%, 10\% 10\% for training, validation and testing.
We measure the accuracy of the attacker model on each test set. 

\paragraph{Utility Metrics}
We evaluate the utility loss when performing obfuscation similarly to the privacy classifier. For each downstream task dataset, we train a DeBERTa model to quantify utility preservation after text obfuscation.
In addition, we also measure the impact on content preservation and soundness (see Appendix~\ref{apx:content-preservation}).

\paragraph{Baselines} 
We use the following baselines:
\subparagraph{Original Text}
 We measure the performance of utility / privacy classifiers when evaluated on original data, the goal of AO would be to decrease the performance of privacy classifiers without decreasing too much the accuracy of utility classifiers.

\subparagraph{Synonyms} As a baseline, we perform a naive text edition using synonyms. We use GPTZzzs\footnote{\url{https://github.com/Declipsonator/GPTZzzs}} to process original texts, it employs a dictionary of synonyms to replace a given proportion of words with their counterparts. The goal of this baseline is to evaluate the attacker behavior when very small edits are made in the original text.

\subparagraph{ALISON} We use ALISON, a recent state-of-the-art text edition AO model leveraging small replacements using a pretrained BERT model. Replacements spans are computed using a threshold on the explanations of an adversarial authorship attribution classifier trained on each evaluation dataset. We train this classifier on each training and validation set before evaluation.

\subparagraph{GPT-3.5} Lastly, we include a comparison with GPT3.5 (\texttt{gpt-3.5-turbo}) \cite{ouyang2022training} as a text generation baseline. We use a simple text obfuscation prompt to capture zero-shot capabilities of GPT-3.5 to perform AO. The prompt used can be found in Appendix~\ref{chat-gpt-prompt}.

\begin{table*}[ht]\centering
    \resizebox{\textwidth}{!}{
\begin{tabular}{l|cccc|cccc|cccc}
\toprule
       & \multicolumn{4}{c|}{IMDb}& \multicolumn{4}{c|}{BAC} & \multicolumn{4}{c}{AMT} \\
Method & \multicolumn{2}{c|}{10 Authors}     & \multicolumn{2}{c|}{20 Authors} & \multicolumn{2}{c|}{10 Authors}     & \multicolumn{2}{c|}{20 Authors} & \multicolumn{2}{c|}{10 Authors}     & \multicolumn{2}{c}{20 Authors} \\
       & Util. $\uparrow$ & \multicolumn{1}{c|}{Attr. $\downarrow$} & Util. $\uparrow$ & Attr. $\downarrow$ & Util. $\uparrow$ & \multicolumn{1}{c|}{Attr. $\downarrow$} & Util. $\uparrow$ & Attr. $\downarrow$ & Util. $\uparrow$ & \multicolumn{1}{c|}{Attr. $\downarrow$} & Util.  $\uparrow$& Attr. $\downarrow$\\ 
       \midrule
Original & 73.51 & \multicolumn{1}{c|}{99.78} & 79.46 & 99.80 & 46.73 & \multicolumn{1}{c|}{61.05}& 53.80 &61.14 &100 & \multicolumn{1}{c|}{70.37}& 86.11 & 42.86\\
Synonyms & 70.38 & \multicolumn{1}{c|}{94.52} & 76.60 & 96.08 & 46.24 & \multicolumn{1}{c|}{59.06}& 51.20 & 58.18 & 91.67 & \multicolumn{1}{c|}{64.81}& 86.11 & 36.90\\
\midrule
 ALISON &61.88 & \multicolumn{1}{c|}{89.59} &\textbf{65.72} &91.02 &\textbf{40.70} & \multicolumn{1}{c|}{40.67}&41.00 &39.22 &\textbf{91.67} & \multicolumn{1}{c|}{70.37}&73.33&35.84 \\
GPT-3.5 &63.33 & \multicolumn{1}{c|}{66.67} &47.37 &35.00 &37.20 &\multicolumn{1}{c|}{42.73 }&\textbf{44.74} &31.27 &60.00 & \multicolumn{1}{l|}{44.44}&61.11 &31.14\\
SFT &\textbf{64.51} & \multicolumn{1}{c|}{62.50} & 39.47&80.00 &40.41 & \multicolumn{1}{c|}{32.44}&40.10 &28.28 &90.00 & \multicolumn{1}{l|}{26.85}&\textbf{75.00 }&21.23\\
TAROT-PPO & 63.54& \multicolumn{1}{c|}{88.89} & 47.37 & 71.67& 35.38& \multicolumn{1}{c|}{29.14}&42.30 &33.62 &90.00 & \multicolumn{1}{c|}{35.19}&72.22 &17.86\\
TAROT-DPO & 57.14& \multicolumn{1}{l|}{\textbf{34.74}} &60.72 &\textbf{17.34} &24.57 & \multicolumn{1}{c|}{\textbf{23.97}}&28.39 &\textbf{16.42} &86.67 & \multicolumn{1}{l|}{\textbf{22.22}}&64.18 &\textbf{16.67}\\
\bottomrule
\end{tabular}}
\caption{Evaluation results (Util: classifier accuracy on utility labels, Attr: authorship attribution accuracy) Best values are \textbf{bolded}.}
\label{tab:privacy-utility-results}
\end{table*}

\subsection{Realistic Attackers}
So far, the considered privacy evaluation adopts the perspective of an attacker who does not have any knowledge about the obfuscation algorithm used. In a more realistic setting, the attacker can likely identify and reproduce the AO model, and perform more advanced attacks by creating adversarial threat models. Following \citet{zhai-etal-2022-adversarial}, we also evaluate our obfuscation models against two enhanced authorship attribution attackers, better suited to simulate real-world attack scenarios. We list the different attackers and their specific aspects based on adversarial training:
\begin{itemize}
    \item \textit{Mix of original and obfuscated texts:} The attacker knows which AO algorithm was used and leverages this knowledge to create a new attribution model. This model is trained on a combination of original source texts and obfuscated texts generated by the known AO algorithm. We use a 50/50 distribution between original and obfuscated data to train this attacker.
    \item \textit{Only obfuscated texts:} While the attacker is also aware of the AO algorithm, they train their authorship attribution classifier exclusively on the obfuscated samples. \citet{zhai-etal-2022-adversarial} demonstrated that this attack setting achieves the highest performance against text edition obfuscations.
\end{itemize}

For each attack scenario, we train a new authorship attribution classifier using the same parameters (see Appendix~\ref{app:hyperparameters} for hyperparameters) and compare the accuracy change from the original attacker.

\subsection{Training new utility models with obfuscated texts}
We experiment with a second use case to evaluate the downstream utility of obfuscated texts. We use the obfuscated texts of each method as a new training set for our utility classifier. This is useful to evaluate each method capability to generate useful training data that can be further used to train a new classifier on the same utility task.

\section{Results}
\paragraph{Downstream Effectiveness}
In Table~\ref{tab:privacy-utility-results}, we present the accuracy change of privacy and utility classifiers. We observe that both SFT, PPO and DPO reduce the attacker accuracy compared to text edition methods (Synonyms and ALISON).
PO helps to learn a good privacy-utility trade-off by largely improving the privacy of obfuscated texts compared to baselines, while preserving similar utility. We observe that DPO consistently outperforms the PPO algorithm on privacy preservation, while using the same base reward model. DPO is also the best-performing privacy preservation over all baselines, with a notable drop of $82,46\%$ on IMDB-20. Note that the utility decrease is larger for the BAC dataset, which could be explained by the number of short texts contained in the dataset, whose edits affect a lot more the end utility. \name-DPO also outperforms GPT-3.5 by providing more utility and less attribution on IMDB-20, AMT-10 and AMT-20. The effectiveness of \name-PPO lays in its utility preservation capabilities. While not being as private, the utility drop is reduced on nearly each dataset compared to \name-DPO. 

\begin{table*}[!htb]
\small
\centering
\begin{tabularx}{\textwidth}{lX}
\toprule
Method & Output \\
\midrule
Original & I loved the whole story even though it was a tad corny at times . I think great acting and the content of the story kept it going. \\ 
\midrule
Synonyms & I loved the quite whole story very even though it was a tad corny at times. I imagine too outstanding playing and the contents of the story kept it sledding. \\
\midrule\midrule
ALISON & I thoroughly enjoyed the entire story even it did have a tad corny at times. I believe the great acting and the story's content were the main reasons to keep it going. \\
\midrule
GPT-3.5 & The entirety of the narrative was quite delightful, despite occasional moments of cheesiness. I believe the stellar performances and the substance of the storyline sustained its momentum. \\
\midrule
SFT & I loved the whole story. It had many good parts and the writing was excellent. I think great acting and the subject matter of the story kept it going.  \\
\midrule
\name-PPO & I loved the whole thing. It was a good story and well-written. It also kept me going at times. I think great acting and the content of the story kept me going. \\
\midrule
\name-DPO & I love the whole story. It's full of action, personality and humour. It keeps me going, though, and the content keeps me going. \\
\bottomrule
\end{tabularx}
\caption{
    Obfuscation example from the IMDb dataset. 
}
\label{tab:ex}
\end{table*}

\begin{figure}[h!]
    \centering
    \includegraphics[width=1\columnwidth]{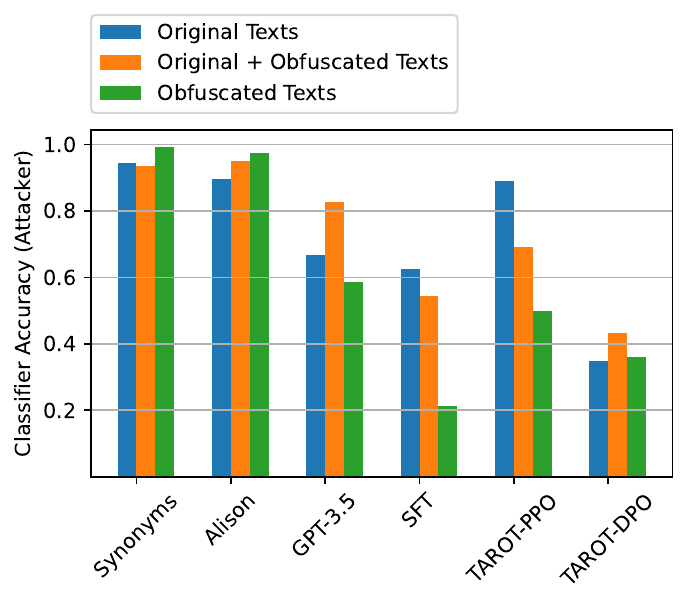}
    \caption{Authorship adversarial training accuracy results on IMDB-10 (lower is better). Generation models are resistant to adversarial training, compared to text edition methods.}
    \label{fig:privacy-retraining}
    \vspace{-4mm}
\end{figure}

\paragraph{Adversarial Attackers}
Figure~\ref{fig:privacy-retraining} highlights the accuracy of adversarial threat models on the IMDb-10 dataset. This attack strategy is effective against text edition approaches (Synonyms and ALISON) as shown by the accuracy gain compared to the base attack only trained on original texts. However, text generation methods (GPT-3.5, SFT, \name-PPO and \name-DPO) show resistance to adversarial threat models, and only GPT-3.5 and \name-DPO are susceptible to the attacker trained on a mix of original and obfuscated texts. This encourages the path of generation methods as promising obfuscators. Note that this is the first obfuscation approach that is shown to be resistant to threat models.\footnote
{\citet{zhai-etal-2022-adversarial} did not include generation models in their study of AO evaluation.}

\begin{figure}[h!]
    \centering
    \includegraphics[width=1\columnwidth]{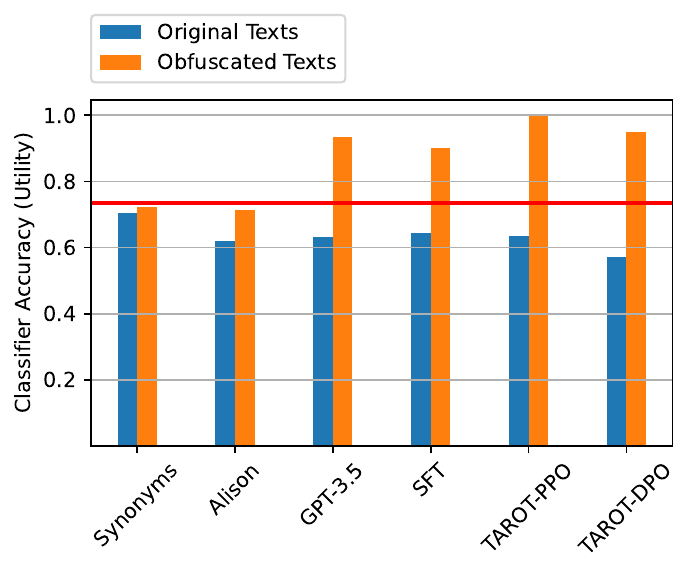}
    \caption{Utility classifier accuracy once trained on IMDB-10 obfuscated texts (higher is better). The {\color{red} red} line indicates the classifier accuracy when trained and evaluated on original data. The overall utility always increases after training on obfuscated texts, this is key to compensate the utility drop of generation methods.}
    \label{fig:utility-retraining}
    \vspace{-4mm}
\end{figure}

\paragraph{Utility Preservation After Retraining}
Figure~\ref{fig:utility-retraining} presents the accuracy of a new utility classifier once trained with obfuscated texts. We observe that the drop in accuracy caused by obfuscation can be compensated by training a new classifier, with an accuracy increase for all methods. Moreover, generation methods are even better candidates for training data, as the final accuracy is higher than the original classifier accuracy. \name-PPO and \name-DPO are the best-performing approaches on this dataset. This highlights the possibility of creating obfuscation methods that are both preserving privacy and keeping utility for training purposes. 

\paragraph{Qualitative Analysis}
We show an obfuscation example in Table~\ref{tab:ex} for each method. The base Synonyms obfuscation results in awkward phrasing and less natural language, compromising readability. ALISON maintains coherence and clarity with slight formalization (``thoroughly enjoyed'' instead of ``loved''). GPT-3.5 significantly rephrases the text using sophisticated language. SFT simplifies and shortens the text, retaining clarity but reducing stylistic nuances. \name-PPO simplifies further, introducing some repetition, which makes the text less formal but still clear. \name-DPO alters the content more significantly, introducing new themes and repetition that can distract from the original meaning. The application of PO assists the text simplification SFT model in making additional modifications to the text. Although these changes in some cases alter the text's meaning, they preserve its overall utility. Appendix~\ref{apx:more_examples} provides more obfuscation examples from proposed and baseline methods.

\paragraph{Ablation Study}
As a complement, we perform an ablation study of each component of our reward model in Appendix~\ref{app:reward-ablation}. It confirms the importance of using a combination of both privacy and utility rewards to learn this trade-off for obfuscation, especially for PPO.

\section{Conclusion}
We introduced a novel authorship obfuscation framework that focuses on optimizing the privacy-utility trade-off for a specific downstream data usage. We fine-tuned a text simplification model using two policy optimization algorithms to obfuscate the authorship of a given text, while preserving utility for multiple tasks. Our end-models are tuned using two sentence embedding rewards, one for content preservation and one for privacy, resulting in an unsupervised approach made for the open-world authorship setting. The results obtained help to improve the privacy from state-of-the-art AO methods, while preserving task utility. Our findings suggest that editing approaches are not suitable for privacy, especially against realistic attack settings. Additionally, we show that generated texts can be used to retrain utility classifiers and increase their performances, while limiting the accuracy of more advanced attackers. Ultimately, the performance of obfuscation methods largely varies depending on the downstream task choice, as does the resulting privacy-utility trade-off, highlighting the importance of selecting an appropriate model based on the specific requirements of the intended application. This calls for more research to design robust evaluation benchmarks for obfuscation systems, to assess and catch failure cases that can map to different real-world scenarios.

\section{Limitations}
The use of LM as text generators for obfuscation is not without risks, LM are known for their hallucination capabilities, so even if the downstream task is not affected, there is still a possibility that the trained LM generated plausible but false text from the original text. As we did not study the content preservation of resulting texts, we do not emphasize the risk of spread of misinformation or harm that can be generated by our fine-tuned LM.

Another limitation of our approach is that we rely on very small language models (380M parameters for \texttt{GPT2-medium}, our SFT baseline), which benefits from limited memory usage but suffers from a restricted context size for generation. As a result, our method tends to reduce the text length, especially for longer texts. This limitation could be mitigated by increasing the size of the SFT model.

Finally, these methods can be limited when applied to short texts, as the replacements create significant changes that directly affect the utility task. 

\section{Ethical Considerations}
In this work, we present authorship obfuscation methods that are intended for beneficial purposes (learning insights from data while preserving privacy). But we recognize that this task presents some risks of misuse.  It can facilitate harmful activities such as posting misinformation, spam, or harmful content, without accountability because of obfuscation. Moreover, these techniques might infringe on intellectual property rights by obscuring the authorship of creative works, depriving creators of their deserved credit. We strongly encourage users to carefully consider these potential dangers before employing such methods.

\bibliography{main}
\bibliographystyle{acl_natbib}

\appendix
\newpage
\newpage
\section{Experimentation Details}
\subsection{Hardware and code}
\label{apx:hardware-code}
We conducted all experiments with Nvidia A30 GPU card with 24GB memory and Intel Xeon Gold 5320 CPU. The main libraries used include \texttt{Pytorch} 2.2.2, Huggingface \texttt{transformers} 4.39.3, \texttt{datasets} 2.19.0, \texttt{tokenizers} 0.15.2, \texttt{trl} 0.8.6, \texttt{evaluate} 0.4.1 and \texttt{sentence-transformers} 3.0.0. Due to memory constraints, models are loaded with \texttt{float16} mixed precision.

Training time for PPO ranges from 15-20 hours, while time for DPO ranges from 6-12 hours. Evaluation time ranges approximately from 19-32 hours.

\subsection{GPT-3.5 prompt}
\label{chat-gpt-prompt}
In our study, we compare with zero-shot prompting using GPT-3.5, a model with approximately 175 billion parameters.  We obfuscate each text on a paragraph level, where the entire text is obfuscated as a unit. We use the following prompt to generate obfuscated texts:
\textit{"Rewrite the following paragraph so that the author’s style is obfuscated."}

\subsection{DPO training}
\label{app:dpo-training}
While both PPO and DPO algorithms methods aim to optimize a model's performance based on a reward function, they differ in their approach to policy optimization. PPO uses a surrogate objective function that approximates the true objective function, while DPO directly optimizes the likelihood of generating a response chosen from a preference dataset over another response. 
This preference dataset is typically collected by having human annotators compare pairs of responses generated by a model and indicate which one is preferred. However, this protocol is impractical for authorship obfuscation because it is difficult to evaluate with human annotations. Therefore, we apply an initial preprocessing step to generate the preference dataset before DPO fine-tuning. We generate preference pairs from SFT outputs, and rank these preferences using the same reward model as PPO. Algorithm~\ref{algo:DPO-dataset} outlines our method for creating this preference dataset for DPO. Preliminary experiments showed that removing samples with closely similar authorship rewards accelerates training convergence. So we specify filtering thresholds $\epsilon_{priv}$ and $\epsilon_{util}$. After testing multiple values, we set $\epsilon_{priv}= 0.10$ and $\epsilon_{util} = 0.05$

\begin{algorithm}%
\caption{Preference Dataset Generation}
\label{algo:DPO-dataset}
   \begin{algorithmic}
   \Require SFT dataset $\mathcal{D}$, privacy threshold $\epsilon_{priv}$, utility threshold $\epsilon_{util}$
   \State prompts = []
   \State chosen = []
   \State rejected = []
        \For {prompt $\in \mathcal{D}$}
           \State left, right = generations from the SFT model
           \State $R_{util-left}$, $R_{priv-left}$ = privacy and utility rewards from the left obfuscation candidate
           \State $R_{util-right}$, $R_{priv-right}$ = privacy and utility rewards from the right    obfuscation candidate
            \If {$\|R_{priv-right}$ - $R_{priv-left}\| >\epsilon_{priv}$ and $\| R_{util-right}$ - $R_{util-left} \| < \epsilon_{util}$}
                \If {$R_{priv-right}$ > $R_{priv-left}$}
                    \State prompt.append(prompt)
                    \State chosen.append(right)
                    \State reject.append(left)
                \Else
                    \State prompt.append(prompt)
                    \State chosen.append(left)
                    \State reject.append(right)
                \EndIf
            \EndIf
        \EndFor
    \State \textbf{return} prompts, chosen, rejected
    \end{algorithmic}
\end{algorithm}

\subsection{Hyperparameters}
\label{app:hyperparameters}
Table~\ref{tab:po-hyperparams} and Table~\ref{tab:classif_params} present hyperparameters used for PO algorithms and evaluation classifiers. Due to limited time and computational resources, we are unable to conduct an exhaustive search across all hyperparameters. Instead, we report the best-performing hyperparameters we identified.

\begin{table}[!ht]
\centering
\resizebox{\columnwidth}{!}{
\begin{tabular}{lcc}
\toprule
               & \name-PPO       & \name-DPO     \\
\midrule\midrule
learning rate  & 1.47e-5   & 2.96e-5  \\
batch size     & 16        & 32    \\
\# epochs      & 3         & 3     \\
KL coef / beta & 0.2       & 0.1   \\
top p          & 1.0       & 1.0   \\
temperature    & 1.0       & 1.0   \\
\bottomrule
\end{tabular}}
\caption{
    Training hyperparameters for PO algorithms.
}
\label{tab:po-hyperparams}
\end{table}

\begin{table}[!ht]
\centering
\resizebox{\columnwidth}{!}{
\begin{tabular}{lcc}
\toprule
               & DeBERTa (Privacy)  &  DeBERTa (Utility) \\
\midrule\midrule
learning rate  & 2e-5   & 2e-5  \\
batch size     & 8        & 8    \\
\# epochs      & 3         & 3     \\
\bottomrule
\end{tabular}}
\caption{
    Training hyperparameters for evaluation models.
}
\label{tab:classif_params}
\end{table}

\subsection{Baseline implementation details}
\subparagraph{Synonyms} We use GPTZzzs to process original texts, it employs a dictionary of synonyms to replace a given proportion of words with their counterparts. The goal of this baseline is to evaluate the attacker behavior when very small edits are made in the original text. We use the \texttt{FinNLP} synonym list and ask the algorithm to change up to 90\% of words, and 80\% of adjectives.

\subparagraph{ALISON} We use the author's code implementation of ALISON, we use the largest edition parameters ($L = 250$ and $c = 1$) to edit the final text as much as possible.

\subparagraph{GPT3.5} We use the \texttt{gpt-3.5-turbo} API endpoint from OpenAI to compute obfuscation, with default temperature, max\_tokens and top\_p.

\section{Content preservation and soundness study}
\label{apx:content-preservation}
We also study the impact on content preservation when obfuscating the text with generation models, including \name. Table~\ref{tab:content-preservation} presents multiple content preservation metrics on the IMDB-10 dataset. Naturally, text edition methods obtain the best content preservation scores, compared to generation methods. In contrast, generation methods are superior in terms of linguistic acceptability (CoLA), since they generate the complete text as a whole. \name-DPO outperforms other methods on this metric.

\begin{table*}[!ht]
\centering
\begin{tabular}{lccccccc}
\toprule
               & Rouge-1 & Rouge-2 & Rouge-L & BLEU & METEOR & BERT Score & CoLA \\
\midrule\midrule
Original  & - & - & - & - & - & - & 69.31  \\
Synonyms  & 83.86 & 68.61 & 83.68 & 64.64 & 92.41 & 94.61 & 30.20  \\
\midrule
ALISON  & 98.24 & 97.08 & 98.19 & 67.48 & 97.61 & 99.01 & 43.88 \\
GPT-3.5  & 38.13 & 11.90 & 29.15 & 6.81 & 33.61 & 81.81 & 73.82 \\
SFT  & 55.69 & 34.04 & 43.20 & 24.06 & 41.13 & 85.58 & 66.66 \\
\name-PPO  & 51.33 & 29.36 & 38.67 & 20.77 & 37.93 & 84.50 & 74.46 \\
\name-DPO  & 42.52 & 17.27 & 29.14 & 10.77 & 30.04 & 80.56 & 81.10 \\
\bottomrule
\end{tabular}
\caption{
    Content preservation scores on the IMDB-10 dataset.
}
\label{tab:content-preservation}
\end{table*}

\section{Complete Evaluation Results}
\label{apx:bac-amt-eval}
\begin{figure*}[h!]
    \centering
    \includegraphics[width=1\linewidth]{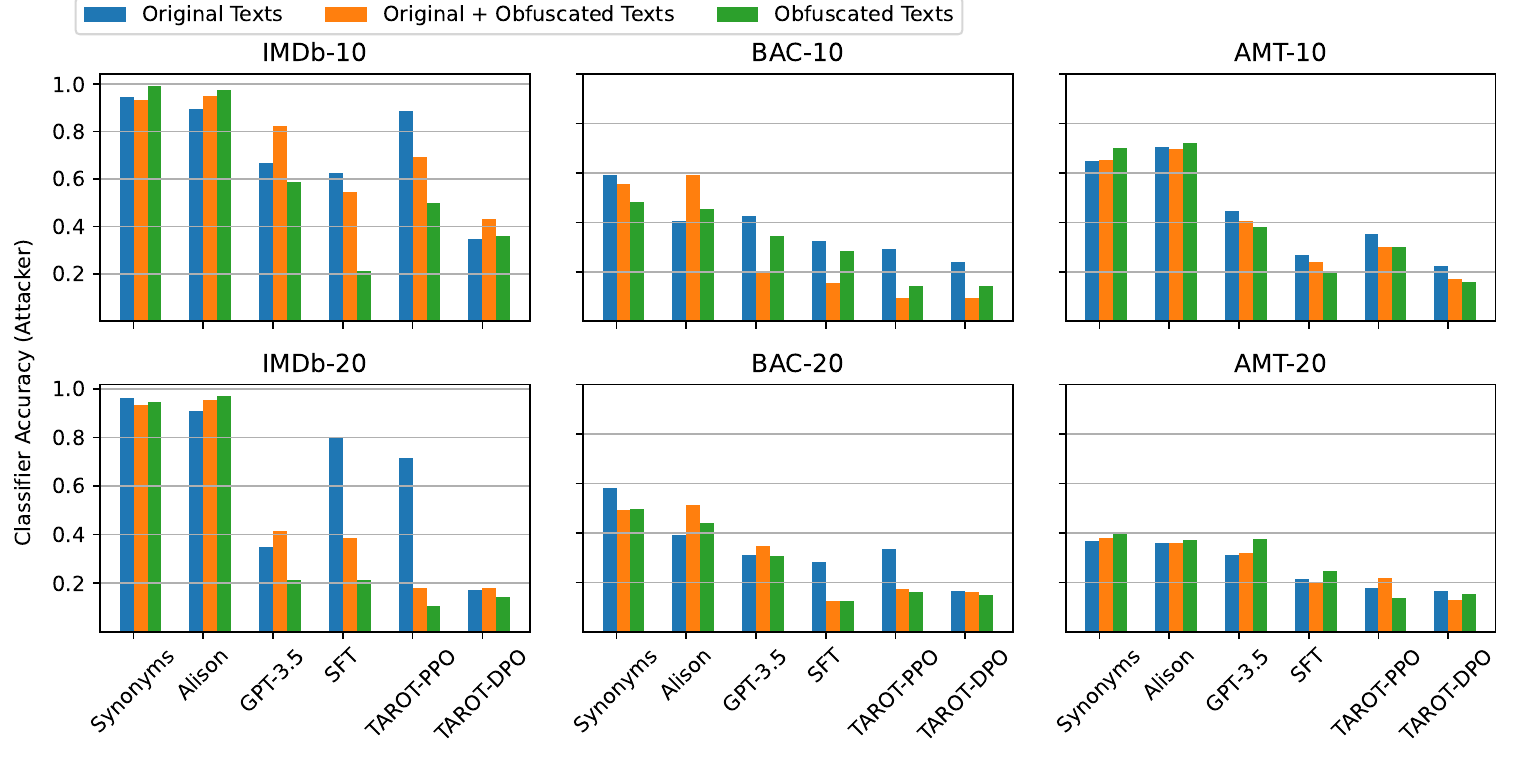}
    \caption{Adversarial training accuracy results (lower is better).}
    \label{fig:privacy-retraining-full}
    \vspace{4mm}
\end{figure*}

Figure~\ref{fig:privacy-retraining-full} presents the complete evaluation results of adversarial training on all datasets.

\begin{figure*}[h!]
    \centering
    \includegraphics[width=1\linewidth]{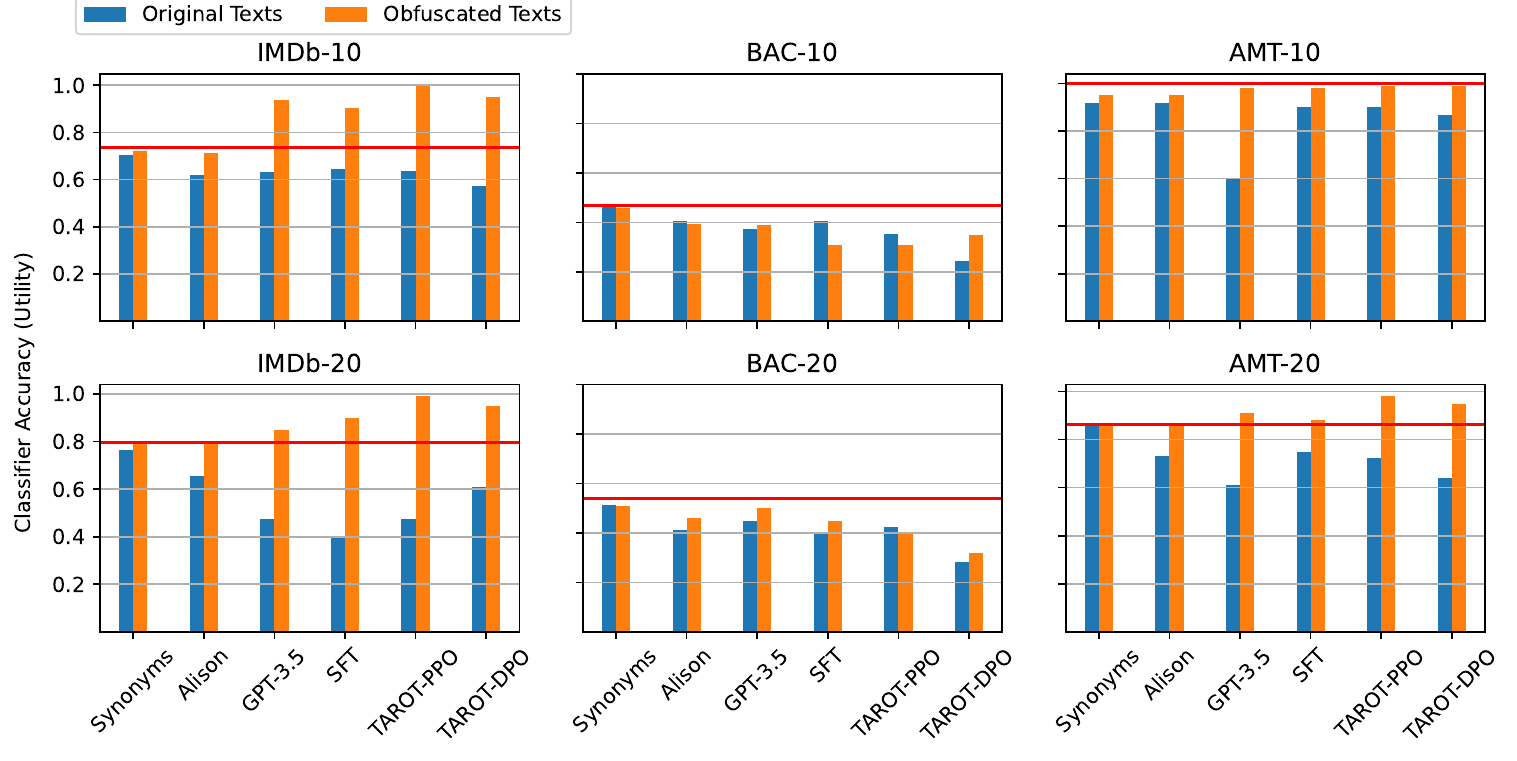}
    \caption{Utility classifier accuracy once trained on obfuscated texts (higher is better). The {\color{red} red} line indicates the classifier accuracy when trained and evaluated on original data.}
    \label{fig:utility-retraining-full}
    \vspace{-4mm}
\end{figure*}

Figure~\ref{fig:utility-retraining-full} presents the complete utility evaluation after retraining on each dataset. The findings presented for IMDb-10 persist for IMDB-20 and AMT-20. 
We observe a smaller change in utility over the AMT-10 dataset due to the high base accuracy of the original classifier (1.0). However, this result does not hold for the BAC-10 and BAC-20 datasets, which is due to the lack of utility preserved after obfuscation. The blog authorship corpus dataset consists mainly of short texts, making it challenging for rewriting methods to transform the text without significantly affecting utility. This issue persists even after retraining the classifier on the obfuscated data.

\section{Reward model ablation study}
\label{app:reward-ablation}
We perform a reward model ablation study to evaluate the importance of each reward component. Table~\ref{tab:ablation} presents the reward value after training on different setups. We observe that the utility preservation and privacy components are both necessary to balance the privacy-utility trade-off. When we remove the LUAR-based reward, it leads to better GTE similarity at the expense of privacy. Similarly, removing the GTE reward leads to better privacy scores at the expense of utility. In practice, removing the privacy reward leads to models that try to copy the original text. While removing the utility reward leads to very short text, with only few words.

\begin{table}[ht]\centering
\resizebox{\columnwidth}{!}{
\begin{tabular}{lcccc}
\toprule
Method & \multicolumn{2}{c}{\name-PPO}     & \multicolumn{2}{c}{\name-DPO} \\
       & LUAR & GTE & LUAR & GTE \\ 
       \midrule
No privacy & 0.975 & 0.993 & 0.983 & 0.977 \\
No utility & 0.403 & 0.421 & 0.706 & 0.633 \\
\midrule
No ablation & 0.931 & 0.825 & 0.915 & 0.738 \\
\bottomrule
\end{tabular}}
\caption{Reward model values when removing one component. A high LUAR value indicates low privacy, and a high GTE value high utility.}
\label{tab:ablation}
\end{table}

\section{Scientific Artifacts}
We list in this section the licenses used in this paper: 
\paragraph{Models}
DeBERTa-v3 (MIT)
Keep It Simple (apache-2.0)
LUAR (apache-2.0)
GTE (apache-2.0)

\paragraph{Software}
GPTZzzs (GPL-3.0)
ALISON (MIT)
GPT-3.5 (Terms of use\footnote{\url{https://openai.com/policies}})
Pytorch (BSD-3)
Huggingface \texttt{transformers}, \texttt{transformers}, \texttt{datasets}, \texttt{trl}, \texttt{evaluate} and \texttt{sentence-transformers} (apache-2.0)

\clearpage
\onecolumn
\section{Additional Obfuscation Examples}

\label{apx:more_examples}

\begin{table*}[!htb]
\small
\centering
\begin{tabularx}{\textwidth}{lX}
\toprule
Dataset: IMDB & \\
\toprule
Method & Output \\
\midrule
Original & Walter Matthau was great as an aging man who lives to hang out in Central Park and tell outrageous lies. He strikes up a relationship with a man whom he drives nuts with his outlandish yarns. Matthau tries a con on several people trying to help them and himself out, but they all backfire . This film had a story that had me riveted to my chair. The entire production was filmed in Central Park, the scenery was beautiful, with many shots of the surrounding buildings. Wonderful movie - 5 stars.\\ 
\midrule
Synonyms & Walter Matthau was really great as an aging man who lives to fall out in Central Park and evidence outrageous lies. He strikes up a relationship with a man whom he drives nuts with his outlandish yarns. Matthau tries a con on several people trying to assist them and himself out, but they all backfire. This film had a story that had me riveted to my chairman. The very full production was filmed in Central Park, the scenery was too beautiful, with extremely many shots of the surrounding buildings. Wonderful movie - 5 stars. \\
\midrule\midrule
ALISON & Walter Matthau was great as an aging man to hang out in central park and tell outrageous lies. he strikes up a relationship with a man whom he drives nuts with his outlandish yarns. Matthau tries a con on several people trying to help them and himself out, but they all backfire. this film had a story that had me riveted to my chair. the entire production vet vet in central park, the scenery was beautiful, with many shots of the surrounding buildings. wonderful movie - 5 stars.\\
\midrule
GPT-3.5 & In portraying a mature chap who thrives on loitering in the heart of Central Park, Walter Matthau delivered an exceptional performance filled with fanciful fables. His character forms a peculiar bond with an individual who becomes perturbed by his fantastical tales. Matthau embarks on a series of deceits aimed at aiding both himself and others, all of which result in unfortunate outcomes. This cinematic masterpiece captivated me, ensuring I remained glued to my seat throughout. The entirety of the picture was shot within the confines of Central Park, showcasing its exquisite landscapes and featuring numerous glimpses of the adjacent edifices. A delightful piece of cinema - a resounding five-star rating.\\
\midrule
SFT &  Walter Matthau was wonderful as an aging man who lives in Central Park. He's a regular at the bar where he tells outrageous stories and tells wild stories. He strikes up a good relationship with a man whom he drives nuts with: his yarns. This film is shot in Central Park, with the backdrop of a long-running feud between Matthau and his friends. Wonderful movie.\\
\midrule
\name-PPO & Walter Matthau was a joy to watch as an aging man who lives out in Central Park. He regularly tells wild stories about how he used to drive around town and tell wild lies. But he strikes up a good relationship with one person who wants to help him out and himself. This film has a lot going for it, with lovely scenery and lots of action. Wonderful film - 5 stars.\\
\midrule
\name-DPO & Walter Matthau is one of those people who can never quite find the right balance between being a good friend and being funny. Though he lives in Central Park, he spends a lot of time telling his friends outrageous stories. This time, though, the relationship is quite straightforward. He strikes up a few conversations with people trying to help them get ahead of the game and, at the same time, he strikes up a relationship with a man who drives him nuts.\\
\bottomrule

\toprule
Dataset: BAC & \\
\toprule
Method & Output \\
\midrule
Original & Is anyone else gonna actually blog? It would be helpful you know! \\ 
\midrule
Synonyms & Is anyone else gonna really blog? It would be very helpful you know! \\
\midrule\midrule
ALISON & Is anyone else gonna actually blog? It would be helpful you know\\
\midrule
GPT-3.5 & Does the inclination exist for any additional individuals to engage in the act of blogging? Such contributions would undoubtedly prove advantageous, do you not concur? \\
\midrule
SFT & Any other person writing about something like this? It would be helpful to know what other people are saying about the matter. \\
\midrule
\name-PPO & Any other person blogging about something? it would be helpful to know who else is going to actually blog? \\
\midrule
\name-DPO & Any other person on the list of potential bloggers? it would be helpful to know what they write about. \\
\bottomrule
\end{tabularx}

\end{table*}

\begin{table*}[!htb]
\small
\centering
\begin{tabularx}{\textwidth}{lX}
\toprule
Dataset: AMT & \\
\toprule
Method & Output \\
\midrule
Original & The interplay between ""new"" and ""old"" ideas, methods, and forms in gothic literature gave it the intrinsic contradictory nature it retains today. The definition of ""gothic"" can never seemed to be agreed upon, the story within a gothic novel still seems to lost track of itself occasionally, the conflict between the emotional effect of terror, and the desire for a logical explanation (science and reason gradually began to replace the supernatural as the driving force behind the horrific) all remain. Neither the ancient nor the modern can be removed from the gothic without sacrificing something essential to the genre. [...] \\ 
\midrule
Synonyms &  The interplay between ""new"" and ""old"" ideas, methods, and forms in Gothic lit gave it the intrinsical contradictory nature it retains today. The definition of ""Gothic"" can never seemed to be agreed upon, the story within a Gothic refreshing relieve seems to missed cartroad of itself occasionally, the conflict between the emotional gist of terror, and the want for a logical account (science and intellect gradually began to replace the occult as the impulsive thrust behind the horrific) all remain. Neither the antediluvian nor the very modern can be removed from the Gothic without sacrificing something essential to the genre. \\
\midrule\midrule
ALISON &  The interplay between ""new"" and ""old"" ideas, methods, and forms in gothic literature gave it the intrinsic contradictory nature it retains today. The definition of ""gothic"" can never seemed to gothic gothic upon, the story within a gothic novel still seems to lost track of itself occasionally, the conflict between the emotional effect of terror, and the desire for a logical explanation (science and reason gradually began to replace the supernatural as the driving force behind the horrific) all remain. neither the ancient nor the modern can neither neither from the gothic without sacrificing something essential to the genre. \\
\midrule
GPT-3.5 & The perplexing interplay betwixt the ""novel"" and the ""antique"" concepulations, methodologies, and configurations in Gothic penmanship hath imbued it with a most incongruous essence that endures unto this day. The elucidation of ""Gothic"" hath forever eluded consensus, the narrative ensconced within a Gothic tome oft doth meander astray, the discord betwixt the emotive impact of dread, and the craving for a reasoned explication (where science and rationale inexorably supplanted the otherworldly as the impelling impetus behind the macabre) doth persevereth. Ne'er can the hoary nor the contemporaneous be excised from the Gothic sans forfeiting somewhat indispensable to the genre. 
\\
\midrule
SFT &  The interplay between new and old ideas is what gives gothic fiction its unique character. Methods and forms of storytelling, such as novel characters and their relationship to nature, give it a contradictory nature. The definition of gothic, however, cannot be agreed upon, the story within a gothic novel still seems to lack logic and reason, with terror still guiding the characters' actions and choices (science and reason gradually replaced the supernatural as the driving force, not just emotions). \\
\midrule
\name-PPO &  the interplay between ""new"" and ""old"" ideas, methods, and forms in gothic literature gave it the intrinsic contradictory nature it still retains today. the definition of ""gothic"" is never clearly agreed upon, the story within a gothic novel still seems to lack a logical foundation, and the desire for a logical explanation often clashes with the character's emotions and mental state. both the ancient and the modern are incompatible with the genre. \\
\midrule
\name-DPO & But the two ideas are not linked by a common underlying logic. The definition of ""gothic"" can never seem to agree with reality. It can only seem to be full of contradictions and inconsistencies, with ideas such as new and old, new and old, about which there is never a clear line between right and wrong. \\
\bottomrule

\toprule
Dataset: BAC & \\
\toprule
Method & Output \\
\midrule
Original & New F-Zero title GBA bound - F-Zero: Falcon Densetsu News for Game Boy Advance at GameSpot  yeah out this week to but overseas based on the anime btw which unfortunately is bad  \\ 
\midrule
Synonyms & New F-Zero claim GBA bound - F-Zero: Falcon Densetsu News for Game Boy Advance at GameSpot  yeah out this week to but abroad based on the anime btw which regrettably is bad  \\
\midrule\midrule
ALISON & New F-Zero title GBA bound - F-Zero: Falcon Densetsu news for game boy advance at GameSpot yeah out this week to but overseas based on the anime btw which unfortunately is bad \\
\midrule
GPT-3.5 & A brand-new installment in the F-Zero franchise is set to debut on the Game Boy Advance, known as F-Zero: Falcon Densetsu. The release is anticipated this week, with availability limited to specific regions tied to its anime adaptation, which has been critically panned. \\
\midrule
SFT & New title is GBA, a reference to falcon - a news service for game boy advance for the past three years. Yeah, out this week, though, overseas based on the anime btw which is just rubbish.  \\
\midrule
\name-PPO & New F-Zero title is a straight up rip-off of GBA. It features falcon news for the game boy advance, which again is crap online. Out this week, though, the anime btw are obviously not very good. \\
\midrule
\name-DPO & An updated F-Zero title, this time with GBA, the title of a news show that the game boy advance on is. Yeah, out this week to news shows like F-Zero but overseas based on the anime, which is bad.    \\
\bottomrule
\end{tabularx}

\end{table*}
\begin{table*}[!htb]
\small
\centering
\begin{tabularx}{\textwidth}{lX}
\toprule
Dataset: AMT & \\
\toprule
Method & Output \\
\midrule
Original & Organisms would have the abilities to move, eat, hunt, and think. These functions would be optimized by genetic algorithms. To create this simulation, there were several steps. The first was to decide upon a programming language. The C++ programming language was chosen for its versatility and large pool of tutorial resources. The next step involved writing pseudo-code, or planning out the program itself. The actual program was broken into several functions. The first function, a global function that provided access to all other parts of the program, generated an arbitrary number of environments and filled them with a random number of food items that had a varying energy level based on the environment.\\ 
\midrule
Synonyms & Organisms would get the abilities to go, eat, hunting, and think. These functions would be optimized by genetical algorithms.
To make this simulation, thither were several steps. The 1st was to decide upon a programing speech. The C++ programing speech was elect for its versatility and really big kitty of tutorial resources. The next stair mired penning pseudo-code, or preparation out the programme itself. The factual program was crushed into several functions. The really first function, a worldwide function that provided approach to all other parts of the program, generated an arbitrary number of environments and filled them with a random figure of food items that had a variable vigor raze based on the environs. \\
\midrule\midrule
ALISON & Organisms would have the abilities to move, eat, hunt, and think. these functions would be optimized by genetic algorithms. To create this simulation, there were several steps. The first was to decide upon a programming language. The C++ programming language for its versatility and large pool of tutorial resources. The next step involved writing pseudo-code, or planning out the program itself. The actual program planning planning into several functions. The first function, a global function that provided access to all other parts of the program, generated an arbitrary number of environments and filled them with a random number of food items that had a varying energy level based on the environment.\\
\midrule
GPT-3.5 & Living beings posses a range of functionalities, such as mobility, sustenance consumption, predatory behaviors, and cognitive processes, that are fine-tuned through complex genetic algorithms.
To initiate the simulation, several procedural stages were undertaken. Initially, a choice was made regarding the selection of a programming language. Eventually, the decision was made to utilize the versatile and well-documented C++ language for its wide range of resources. Subsequently, the process involved the formulation of pseudo-code, a strategic blueprint for the development of the program. The program itself was compartmentalized into multiple functions. The primary function, acting as a global facilitator, was designed to generate artificial environments and populate them with random assortments of food items possessing varying energy levels depending upon the environment. \\
\midrule
SFT &  These organisms would have the ability to move, eat, hunt and think. These functions would be better optimized by genetic algorithms. These algorithms would be able to predict what environment a given creature would encounter next, and so on, generating different environments for each task. To create this simulation, programming language was used. Programming languages are usually designed to allow flexible programming of tasks for different stages in a person's life, like diet and training. These environments are then tested out in a variety of ways, like hunting or training. Tucking these algorithms into the program also meant creating more environment specific.\\
\midrule
\name-PPO &  These organisms would have the ability to move, eat, hunt and think. these functions would be optimized by genetic algorithms. These functions would be triggered by training programs that specifically targeted specific traits, such as hunger and thirst. To create this simulation, these would include programming languages for different types of environments, such as languages for animals and plants. These training programs would be tested against real-world situations and filled with different food types.\\
\midrule
\name-DPO &  These organisms are capable of thinking and eating. They can move and hunt, and they have even been observed to think about their environment. These functions, which would normally be optimized by a computer, are now well-suited for creating life on the planet. To train these organisms to perform these functions, for example, is to have a programming language that can be easily adapted to new situations. To put together a simulation of a particular environment, it would need to provide for training a range of animals, from which to choose. These kinds of simulations, which are relatively simple, would be ideal for training mental processes.\\
\bottomrule
\end{tabularx}

\end{table*}

\begin{table*}[!htb]
\small
\centering
\begin{tabularx}{\textwidth}{lX}
\toprule
Dataset: IMDB & \\
\toprule
Method & Output \\
\midrule
Original & Funny romance / comedy with some well written lines for everybody . I especially enjoyed the give and take between Tony Randall and the Rock . Doris Day is soooooo cute ; I had a major crush on her during my junior high years . I liked Lynde's character , but I prefer his more sarcastic roles . 3 stars . \\ 
\midrule
Synonyms & Funny latin / comedy with some swell written lines for everybody . I especially enjoyed the dedicate and take 'tween Tony Randall and the Rock . Doris Day is soooooo very cute ; I had a major smash on her during my junior so high years . I liked Lynde's character , but I favour his more sarcastic roles . 3 stars \\
\midrule\midrule
ALISON &  Funny romance / comedy with some well written lines for everybody. I especially enjoyed the give and take between Tony Randall and the rock. Doris Day is soooooo cute; I had a major crush on her during my junior high years. I liked Lynde's character, but I prefer his more sarcastic roles. 3 stars\\
\midrule
GPT-3.5 &  A delightful tale of amusement and affection with a sprinkling of finely crafted dialogue that will appeal to all. The banter shared between the one called Tony Randall and the individual known as the Rock was particularly engaging. Doris Day exuded a charm that was utterly endearing; during my formative years, a deep infatuation for her blossomed. The portrayal by Lynde was rather enjoyable, though I find myself partial to his portrayals imbued with a sense of sarcasm. A rating of three stars is the verdict bestowed upon this film.\\
\midrule
SFT &  Funny romance / comedy with some funny lines. I loved the give-and-take between Tony Randall and the Rock, which was great to watch as I got older. Doris Day is soooooo cute ; she's always been this way, and I like her more in my junior high years. 3 stars.\\
\midrule
\name-PPO &  Funny romance / comedy with some well-written lines for everybody. I really enjoyed the give-and-take between Tony Randall and Doris Day. I loved her personality and her reaction to situations. They're both cute, but not as smart as Tony Randall. 3 stars.\\
\midrule
\name-DPO &  Funny romance. It's a good combination of comedy and give-and-take. There's always something interesting to say in front of the camera, or in front of the audience, about how they feel about someone else. Doris Day is soo cute, and soo funny. She's very much like me at my younger years, when I played sport. 3 stars."\\
\bottomrule
\end{tabularx}
\caption{
    Additional qualitative examples for each obfuscation method.
}
\label{tab:ex_apx}
\end{table*}

\end{document}